# Comparative Analysis of Deep Learning Models for Brand Logo Classification in Real-World Scenarios

Qimao Yang
Department of ECE
University of Florida
Gainesville, US
qimao.yang@ufl.edu

Huili Chen
Department of Pharmaceutics
University of Florida
Orlando, US
huilichen@ufl.edu

Qiwei Dong
Department of ECE
University of Florida
Gainesville, US
dongqiwei@ufl.edu

*Abstract*—In this report, we present a comprehensive study on the performance of various deep learning models for brand logo classification in real-world scenarios. The dataset consists of $3,717$ labeled images of logos from ten prominent brands, including Nike, Adidas, Ford, Honda, General Mills, Unilever, McDonald's, KFC, Gator, and 3M. Each image has a resolution of $300 \times 300 \times 3$ (RGB). $80$ percent of the data was split into the training set, and the remaining $20$ percent into the validation set.

We investigated the performance of two kinds of deep learning models, including Convolutional Neural Network (CNN) models and Vision Transformer (ViT) model. These models were chosen for their computational efficiency and high performance. The classification results demonstrated that the ViT model, DaViT small, achieved the highest accuracy of $99.60\%$. With a single NVIDIA A100, the DenseNet29 achieved the fastest inference speed of $366.62$ FPS, while the inference time of DaViT small is only $66.86$ FPS. Our findings suggest that the DaViT model, among the models we tried, is the most suitable choice, considering its superior performance in terms of accuracy, for those offline applications. This study provides practical examples of the application of deep learning techniques in real-world brand logo classification tasks.

## I. INTRODUCTION

Brand logo recognition plays a critical role in modern marketing and advertising strategies. However, with the ongoing rapid growth of online and offline media, relying solely on human labor to fulfill the computational needs of implementing these strategies has become insufficient. Consequently, the need for tools that can accurately identify and classify logos in real-world scenarios has become increasingly important. Fortunately, machine learning techniques, particularly deep learning, have demonstrated great potential in addressing these tasks.

Moreover, image classification holds significant meaning for computer vision techniques. Beyond its own applications, image classification can be used as a foundation for various tasks such as object detection [1], semantic segmentation [2], and even multi-modal tasks [3].

The main objective of this report is to evaluate and compare the performance of various state-of-the-art deep learning models for classification in real-world scenarios. By analyzing the performance of different models, our goal is to determine the trade-offs between models for this task. Ultimately, we hope to provide insights into potential solutions for applying image recognition and classification tasks in more complex environments.

In the field of classification, Convolutional Neural Networks (CNNs) have become the most popular models since the emergence of LeNet [4] and AlexNet [5]. Nowadays, DenseNet [6] and MobileNet [7] are widely used in industries for classification tasks and as backbones for other models. DenseNet's dense connectivity pattern allows it to improve gradient flow while maintaining a smaller network structure with comparable performance to other models. Similarly, MobileNetV3, another CNN model introduced in 2013, aims to achieve a balance between accuracy and computation efficiency.

In 2017, the Transformer structure [8] was introduced, increasing the number of model parameters and demonstrating a significant advantage in natural language processing. In recent years, transformer networks have become increasingly popular in various areas, including computer vision. The attention mechanism enables it to model long-range dependencies and global context in images more effectively by directly capturing relationships between all pairs of pixels, regardless of their distance. Among the variants of transformer networks, Dual Attention Vision Transformers (DaViT) [9] caught our attention with one of the best performances ever seen.

The scope of this study encompasses four selected deep learning models (DenseNet29, MobileNetV3 small, MobileNetV3 large, and DaViT small) and the chosen logo dataset, which comprises images taken in various real-world scenarios for ten prominent brands. Although our findings provide valuable insights into the performance of these models, further research is necessary to generalize our results to other brand logo recognition tasks and to explore the potential of other deep learning models.

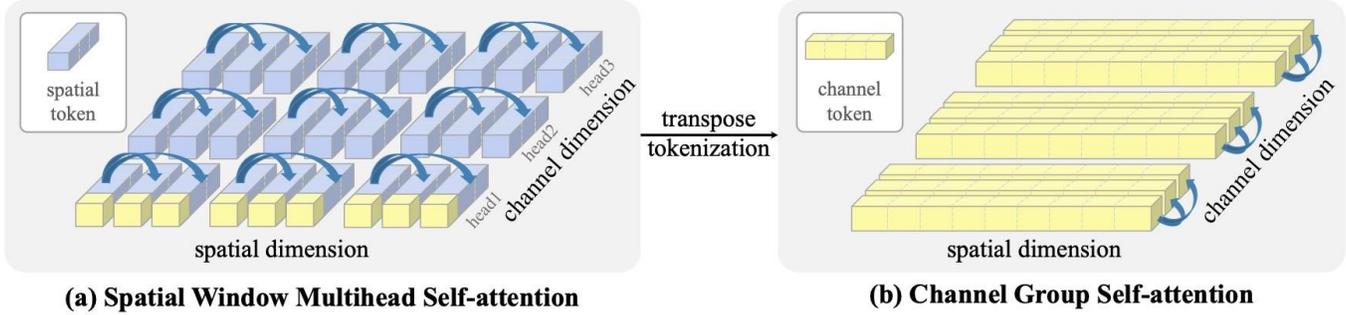

Fig. 1: Two Types of attention modules from [9]: (a) Spatial window multihead self-attention. (b) Channel group single-head self-attention

## II. IMPLEMENTATION

The attention module is a mechanism in transformers that allows models to selectively focus on specific parts of the input data, thereby improving their ability to capture and process relevant information and context for better performance in various tasks. In the case of Vision Transformers, the input images are divided into fixed-size non-overlapping patches, which are then flattened and processed using a transformer-based architecture.

DaViT proposes a dual attention mechanism to efficiently achieve global modeling. The dual attention here refers to self-attention from two perspectives: one is the self-attention of spatial tokens, where the spatial dimensions ($H \times W$) define the number of tokens, and the channel dimensions ($C$) define the feature size of the tokens, which is also the most commonly adopted method in ViT. The other is the self-attention of channel tokens, with the channel dimensions ($C$) defining the number of tokens, and the spatial dimensions ($H \times W$) defining the feature size of the tokens. To reduce computation, both self-attention methods use grouped attention: for spatial tokens, the spatial dimensions are divided into different windows, which the paper refers to as spatial window attention; for channel tokens, similarly, the channel dimensions can be divided into different groups, which the paper refers to as channel group attention. These two types of attention are shown in Figure 1.

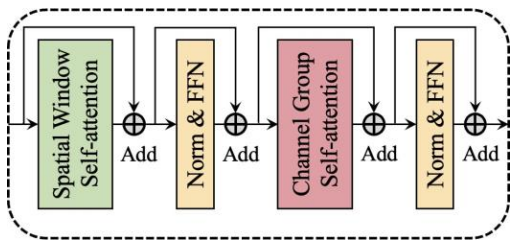

Fig. 2: Dual Attention Blocks in DaViT.

The spatial window attention is capable of extracting local features within windows, while channel group attention can learn global features, as each channel token is global in the image space. DaViT adopts a pyramid structure consisting of four stages, each of which contains a certain number of dual attention blocks. These blocks stack the two types of attention (along with the FFN) alternately, as illustrated in Figure 2.

|  | Output Size | Layer Name | DaViT-Small |
|---|---|---|---|
| stage 1 | 75 × 75 | Patch Embedding | kernel 7, stride 4, pad 3, $C^1$ = 96 |
| | 75 × 75 | Dual Transformer Block | $\left[\begin{array}{c} \text{win. sz. } 7 \times 7, P_w = 49 \\ N_h^1 = N^{1'} = 3 \\ C_h^1 = C_g^{1^g} = 32 \end{array}\right] \times 1$ |
| stage 2 | 38 × 38 | Patch Embedding | kernel 2, stride 2, pad 0, $C^2$ = 192 |
| | 38 × 38 | Dual Transformer Block | $\left[\begin{array}{c} \text{win. sz. } 7 \times 7, P_w = 49 \\ N_h^1 = N^{1'} = 3 \\ C_h^1 = C_g^{1^g} = 32 \end{array}\right] \times 1$ |
| stage 3 | 19 × 19 | Patch Embedding | kernel 2, stride 2, pad 0, $C^3$ = 384 |
| | 19 × 19 | Dual Transformer Block | $\left[\begin{array}{c} \text{win. sz. } 7 \times 7, P_w = 49 \\ N_h^1 = N^{1'} = 3 \\ C_h^1 = C_g^{1^g} = 32 \end{array}\right] \times 1$ |
| stage 4 | 10 × 10 | Patch Embedding | kernel 2, stride 2, pad 0, $C^4$ = 768 |
| | 10 × 10 | Dual Transformer Block | $\left[\begin{array}{c} \text{win. sz. } 7 \times 7, P_w = 49 \\ N_h^1 = N^{1'} = 3 \\ C_h^1 = C_g^{1^g} = 32 \end{array}\right] \times 1$ |

TABLE I: The architecture of DaViT used in this report: $C$: Number of total channels; $P_W$: Patches per window; $N_h$: Number of heads; $N_g$: Number of channel groups; $C_h$: Channels per head; $C_g$: Channels per group.

For our experiments, we utilized the small DaViT structure, which first employs a patch embedding layer, using a 7 × 7 conv with a stride of 4, followed by four stages. The configurations of each stage can be found in Table I.

To optimize our experiments, we used AdamW as the optimizer, and we adopted a 5-epoch warmup of learning rate. Given that we used mixup in our data augmentation, we used the SoftCrossEntropy as the model's loss function.

## III. EXPERIMENTS

### 3.1. Datasets

We conducted a comprehensive experiment to evaluate the performance of various deep-learning models for brand logo classification using our dataset of 3,717 labeled images of logos from ten brands, including Nike, Adidas, Ford, Honda, General Mills, Unilever, McDonald's, KFC, Gator, and 3M.

**Data Preprocessing:** We took multiple steps to prepare the data for training and utilized various techniques to standardize the images for the model. Additionally, we performed data augmentation to increase the dataset's diversity artificially, thereby improving the model's generalization ability. We also noticed that some logo features in the training dataset were small in resolution, which could lead to a loss of information through compression. Therefore, we decided to use the original resolution of $300 \times 300$ on the images to avoid information loss.

**Data Cleaning:** After converting the data from *.npy into images, we found some mislabeled images in the dataset. We corrected these labels and then split the dataset into training and validation sets, with 80% of the data placed into the training set and the remaining 20% into the validation set. In total, we have around 2,957 training images and 760 validation images. While performing label correction, we discovered some images that were challenging to recognize. We tagged these images and placed them in the validation set to evaluate the generalization ability of our models. The challenges presented by the dataset are demonstrated in the examples shown in Figure 3.

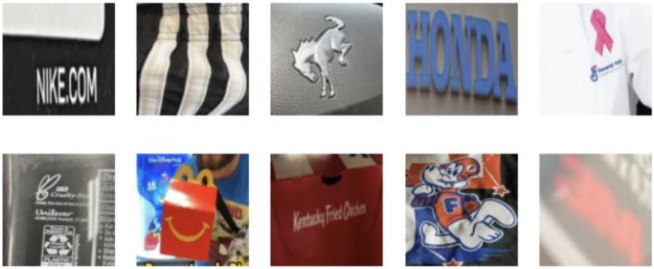

Fig. 3: Examples of images that are difficult to classify for each label.

**Data Augmentation** Data augmentation is an important step in the process of training neural networks for computer vision. It enhances the variability of data and helps to restrain the overfitting problem of the model. Several techniques were applied to augment the training dataset in our study.

- **Classic augmentation methods** like random scale and crop, random flip, random hue, saturation, exposure, and brightness tweak.
- **Mixup[10]**, a domain-agnostic data augmentation technique proposed by Zhang et al., which mixes the features and corresponding labels from two images as follows:

$$\tilde{x} = \lambda x_i + (1 - \lambda)x_j,$$
$$\tilde{y} = \lambda y_i + (1 - \lambda)y_j,$$

  where $x_i$ and $x_j$ are raw input vectors, $y_i$ and $y_j$ are one-hot label encodings, and $\lambda \in [0, 1]$.
- **AutoAugment[11]**, an automatic augmentation policy that searches for the best augmentation policies in data.

However, rather than conducting an online search with the brand logo dataset, we used the searched ImageNet policies with various tweaks and improvements.

### 3.2. Training

**Initial Training:** We initially selected three models, DenseNet29, MobileNetV3 small, and MobileNetV3 large, to represent CNNs. These models were trained from scratch without any fine-tuning, and the initial results were observed.

|  | DenseNet29 | MobileNetV3 small | MobileNetV3 large |
|---|---|---|---|
| # of Params | 174,606 | 1,528,106 | 4,214,842 |
| $Acc_{val}$ | 87.89% | 88.68% | 92.63% |

TABLE II: Validation accuracy of CNN models and their corresponding number of parameters.

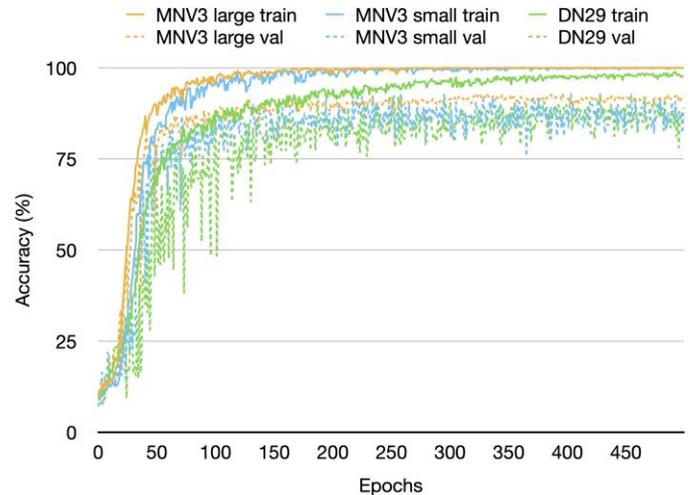

Fig. 4: Training and validation accuracy curves for CNN models.

As shown in Figure 4 and Table II, MobileNetV3 large achieved the best validation accuracy of 92.63%, while the accuracy on the training set was almost 100%. This suggests that the generalization of MobileNetV3 large is not good enough to handle this task. Furthermore, the results indicated that performance improves as the model's size increases, which led us to question the performance of even larger models. We used DaViT small, a transformer model with $48.98M$ parameters, for this purpose. Training such a model from scratch would be challenging, so we used pre-trained weights to initialize our model, except for the output layer due to the mismatched class number. The DaViT small achieved the best validation accuracy of 98.82% without any fine-tuning, indicating that it may be a more effective approach for our dataset compared to CNNs.

**Performance Finetune:** In this section, we illustrate how we obtained our final model and present the results of our experiments.

We evaluated the initial DaViT small model on our validation set using a threshold of 0.5 and inspected the wrongly classified images. Some examples are shown in Figure 5.

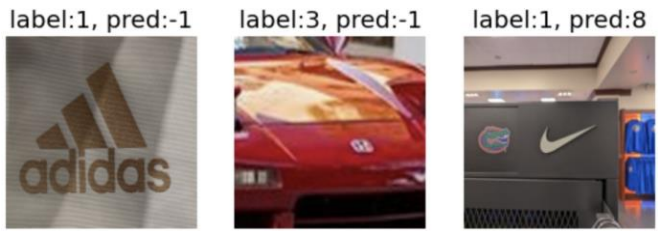

Fig. 5: Examples of incorrectly predicted samples in the validation set.

The mistakes observed in the validation set were classified into three types: (a) errors that should be corrected, (b) errors that could be allowed, and (c) cases that were not mistakes. These three types correspond to the figures in Figure 5. Type (b) refers to those errors that are difficult to recognize even with human eyes, while type (c) corresponds to those images with overlapped features. To improve the model's performance, we focused on correcting the type (a) errors. We identified similar training images with features similar to misclassified images and increased the likelihood of choosing them in the dataloader. We also used the best-performing model from the initial training to initialize the weights and trained it for 100 epochs to evaluate its performance. The final model's validation accuracy improved to 99.60%, as shown in Table III.

|  | Init $Acc_{val}$ | Final $Acc_{val}$ |
|---|---|---|
| DaViT small | 98.82% | **99.60%** |

TABLE III: Validation accuracy comparison of DaViT small before and after finetuning.

We also evaluated the inference speed of the models, and the results are summarized in Table IV. The evaluation was conducted on a single NVIDIA A100 GPU using CUDA 11.7, PyTorch 2.0.0, and FP32.

|  | DenseNet29 | MobileNetV3 small | MobileNetV3 large | DaViT small |
|---|---|---|---|---|
| FPS | **366.62** | 212.71 | 179.39 | 66.86 |

TABLE IV: Inference speed comparison of evaluated models.

Based on our findings, the DaViT model is the most suitable choice for offline applications, given its superior accuracy performance. However, if speed is a concern, the DenseNet29 or MobileNetV3 small would be good options.

## IV. CONCLUSIONS

Our experiments demonstrate that transformer-based deep learning models, such as DaViT, outperform traditional CNNs in image classification tasks in terms of accuracy. Although these models currently have high computational demands, ongoing research focuses on improving their efficiency, making them more practical for a wider range of applications.

While CNNs still have an advantage in real-time responsiveness and privacy constraints, the advancement of computing capabilities and optimization of ViT architectures indicate a promising future for transformer-based models in deep learning.

In summary, although challenges related to computational efficiency persist, the potential of ViT and similar models is significant. As research progresses, we can expect to see a growing impact of transformer-based models in various real-world applications.